\documentclass[conference]{IEEEtran}
\IEEEoverridecommandlockouts

\usepackage{cite}
\usepackage{amsmath,amssymb,amsfonts}
\usepackage{algorithmic}
\usepackage{graphicx}
\usepackage{textcomp}
\usepackage{xcolor}
\usepackage{booktabs}
\usepackage{pifont}
\usepackage{amssymb}
\usepackage{color}
\usepackage{makecell}
\usepackage{multirow}
\def\BibTeX{{\rm B\kern-.05em{\sc i\kern-.025em b}\kern-.08em
    T\kern-.1667em\lower.7ex\hbox{E}\kern-.125emX}}
\begin{document}

\title{4D-CAT: Synthesis of 4D Coronary Artery Trees from Systole and Diastole\\
}

\author{
	\IEEEauthorblockN{
		Daosong Hu\textsuperscript{1†}, 
		Ruomeng Wang\textsuperscript{2,3†}, 
		Liang Zhao\textsuperscript{2†}, 
		Mingyue Cui\textsuperscript{1},	
		Song Ding\textsuperscript{2,3*}
		and Kai Huang\textsuperscript{1*}} 
	\IEEEauthorblockA{\textsuperscript{1}School of Computer Science and Engineering, Sun Yat-sen University}
	\IEEEauthorblockA{\textsuperscript{2}Ren Ji Hospital, Shanghai Jiao Tong University School of Medicine}
	\IEEEauthorblockA{\textsuperscript{3}Punan Branch of Renji Hospital, Shanghai Jiao Tong University School of Medicine}
	\IEEEauthorblockN{†These authors contribute equally.}
	\IEEEauthorblockN{*Email: dingsong1105@163.com (S.D.), huangk36@mail.sysu.edu.cn (K.H.)}
} 

\maketitle

\begin{abstract}
	The three-dimensional vascular model reconstructed from CT images is widely used in medical diagnosis. At different phases, the beating of the heart can cause deformation of vessels, resulting in different vascular imaging states and false positive diagnostic results. The 4D model can simulate a complete cardiac cycle. Due to the dose limitation of contrast agent injection in patients, it is valuable to synthesize a 4D coronary artery trees through finite phases imaging. In this paper, we propose a method for generating a 4D coronary artery trees, which maps the systole to the diastole through deformation field prediction, interpolates on the timeline, and the motion trajectory of points are obtained. Specifically, the centerline is used to represent vessels and to infer deformation fields using cube-based sorting and neural networks. Adjacent vessel points are aggregated and interpolated based on the deformation field of the centerline point to obtain displacement vectors of different phases. Finally, the proposed method is validated through experiments to achieve the registration of non-rigid vascular points and the generation of 4D coronary trees.
\end{abstract}

\section{Introduction}
Coronary artery disease (CAD) is a prevalent and potentially life-threatening condition. The treatment approach involves periodic examination, risk assessment, and timely intervention \cite{garavand2023towards}. Computed Tomography (CT), a non-invasive diagnostic technique for evaluating stenosis, is extensively employed in screening for CAD \cite{xu2023coronary}. The advancement of deep learning technology enables the reconstruction of highly accurate three-dimensional vascular and organ models through neural networks, facilitating an intuitive presentation of lesion locations and aiding in diagnosis \cite{fu2020rapid}. Nonetheless, the coronary arteries exhibit motion, and relying solely on static models from a single phase may result in false positives.

The forceful contraction of the heart induces compression and deformation of the surrounding coronary arteries, resulting in a relocation of the lesion. When compared to static models, dynamic models prove more effective in illustrating the location and morphological changes of stenosis. The point cloud serves as a critical representation for visualizing human organ structures.

4D computed tomography (4D-CT) introduces a temporal dimension to the 3D model. Fu et al. \cite{fu2020lungregnet} introduced an unsupervised deep learning method to predict lung movement, generating intermediate volumes through lung registration at various stages. Owing to their high-quality generation results, Kim et al. \cite{kim2022diffusion} incorporated diffusion models into temporal medical image generation tasks. This method deduces the deformation information of systolic and diastolic phases in cardiac volume data, generating time frames along continuous trajectories. Nevertheless, direct application of these methods to coronary arteries is challenging. Due to the potential harm caused by contrast agents to patients, collecting large-scale continuous data is not feasible. Simultaneously, the coronary artery represents only a small portion in CCTA images. Directly inserting frames by using CCTA images can lead to imbalance issues. Thus, acquiring the intermediate volume directly through coronary artery trees during systole and diastole poses a challenge.


In this paper, we explore a new direction of 4D medical models by generating 4D coronary artery trees using systolic and diastolic point clouds. The proposed method is designed to utilize a two-stage point cloud for non-rigid registration and to acquire a deformation field. The beating process of coronary arteries with the heart is linearly interpolatable over the timeline. Interpolating the predicted deformation field allows the inference of the intermediate volume. Due to ethics, the amount of paired data is limited, making it difficult to directly predict deformation fields through neural networks. Consequently, the centerline is employed to represent blood vessels, and the proposed segmentation, sorting, and deep learning strategies are utilized to obtain the deformation field of blood vessels.

\begin{figure*}[t]
	\centerline{\includegraphics[width=0.96\linewidth, scale=1.00]{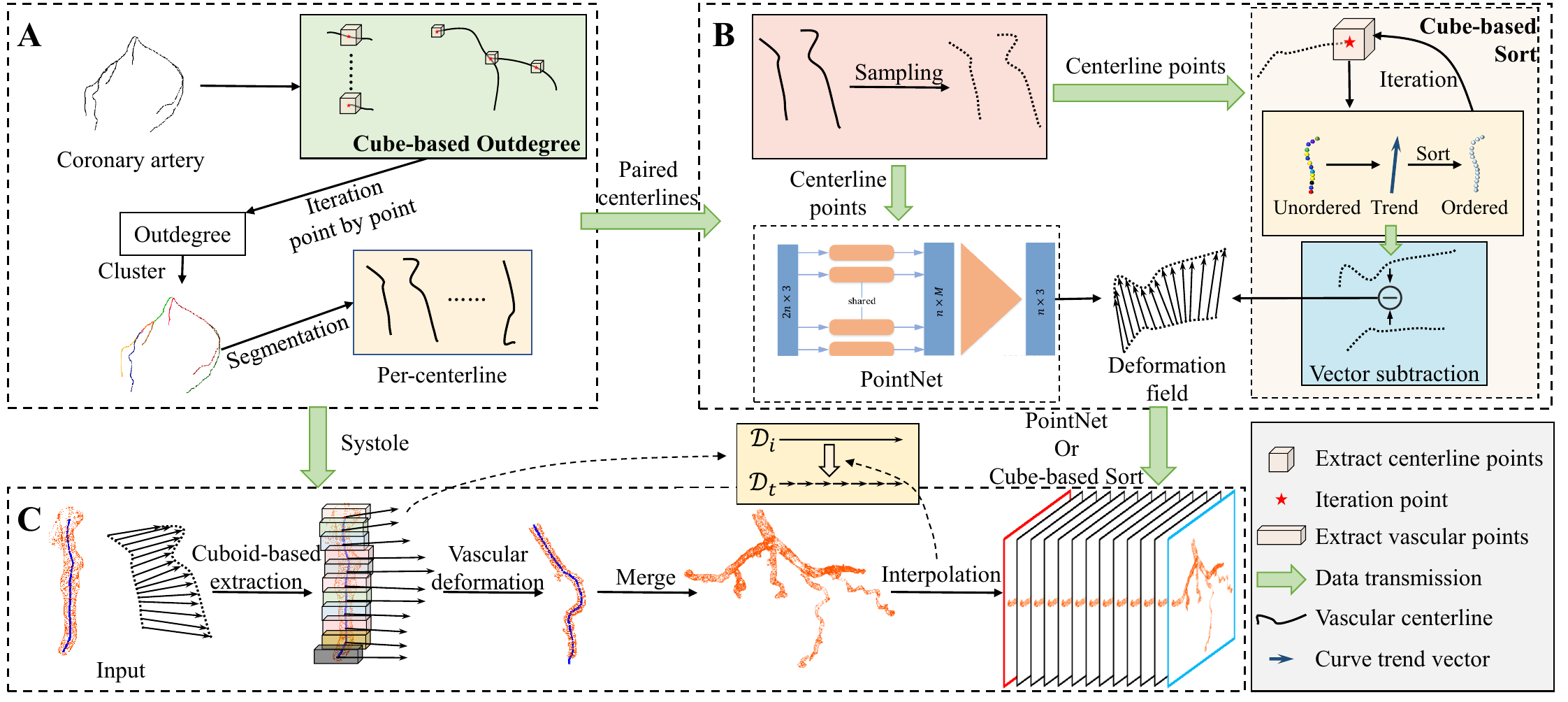}}
	\caption{The overview of our proposed method.}
	\label{Method}
\end{figure*}

\section{Proposed method}

\subsection{Centerline segmentation}
The centerline is a collection of scattered points in three-dimensional space, with certain line characteristics, such as close coordinate values of neighboring points. According to the number of neighboring points owned by different points, these points can be named starting points ($s$), middle points ($m$), and branching points ($b$). As shown in Fig. \ref{Method} (A), we propose a cube-based method that calculates the distance between neighboring points and the surface of the cube to determine whether they intersect. We define the number of intersecting faces as degrees, as follows:
\begin{gather}
	C_{x} = Cube(x, C) \\
	O(x) = \sum_{i=0}^{6} \| C_{x} - c_i \|<\epsilon
\end{gather}
where, $x$ represents the iteration point, and $C$ is the centerline point of vessel. $C_{x}$ denotes the set of points extracted by the cube centered on $x$. $O(\cdot)$ is the outdegree, and $c_i$ represents the six faces of the cube. $\epsilon$ is the threshold, and if the distance is less than $\epsilon$, which is considered that the centerline intersects with this side, and the outdegree is increased by 1. According to the outdegree, the attributes of the centerline point can be defined as:
\begin{equation}
	\mathcal{A}(x) = \left\{
	\begin{aligned}
		s & , & O(x)=1 & \; and \; d(x, \bar{c})=l/2\\
		m & , & O(x)=2 &\\
		b & , & O(x)>2 &\\
	\end{aligned}
	\right.
\end{equation}
where, $\mathcal{A}$ represents the attribute of point $x$. $d(\cdot)$ is the distance from a point to a plane, $\bar{c}$ denotes the plane that does not intersect with the centerline, and $l$ is the length of the cube. During the iteration process, if $\mathcal{A}$ is the starting or middle point, the iteration point $x$ belongs to the same segment as $C_{x}$. If $\mathcal{A}(x)=b$, the point is labeled. If $C_{x}$ without classification or only contains branching points, increase the number of classification.

\subsection{Deformation field prediction}
Point by point deformation field prediction can indicate the displacement path of points. For non-rigid registration of coronary arteries, registration $\mathcal{R}_s$ is achieved by coupling the systolic points $V_s$ and deformation field $\mathcal{D}$. As follows:
\begin{equation}\label{R}
	\mathcal{R}_s = V_s + \mathcal{D}
\end{equation}
Due to the disorder of point clouds, it is difficult to directly obtain the displacement field. The centerline is a representation of the coronary artery, which not only has curved features but also reduces the number of points. As depicted in Fig. \ref{Method} (B), we propose two methods, namely cube-based sort and deep-learning method.
\subsubsection{Deep-learning method}
We propose a network framework based on PointNet \cite{qi2017pointnet}, which connects the centerline of the systolic and diastolic phases and extracts the corresponding relationship between the two centerline points through one-dimensional convolution with shared parameters. Finally, MLP is used to obtain the deformation field. The network's work is as follow:
\begin{equation}
	\mathcal{D}_c = f_{\theta} ([C_s, C_d])
\end{equation}
where $\mathcal{D}_c$ is the deformation field of the centerline, and $\mathcal{D}_c \in \mathbb{R}^{n \times 3}$. $f_{\theta}$ represents the network. $[C_s, C_d]$ is the input, formed by concatenating the centerline of the systolic ($C_s$) and diastolic phases ($C_d$), and $[C_s, C_d] \in \mathbb{R}^{2n \times 3}$. The disorder and independence of point clouds make it difficult to design loss functions. Based on the line features retained in $C_s$ and $C_d$, soft-DTW \cite{cuturi2017soft} is used as the loss function in this task. By searching for the shortest path in two point cloud matrices as the loss function, the details are as follows:

\begin{gather}
	\mathcal{L}_{DTW}(C_d, C_{t}) = -\gamma \log \sum_{r \in R_{\Sigma}} e^{- \langle r, d(C_{d}, C_{t}) \rangle / \gamma} \\
	C_{d} = C_s + \mathcal{D}_c\\
	d(C_{d}, C_{t}) = \sum \| C_{d i}-C_{t j} \|_2^2
\end{gather}
where $\mathcal{L}_{DTW}$ is the soft-DTW, and $\gamma$ is the smoothing parameter. $R_{\Sigma}$ represents all the possible paths. $d$ denotes the Euclidean distance matrix, which includes the Euclidean distances of all corresponding points in the systole and diastole. $C_{d i}$ is the mapped systolic point, and $C_{t j}$ is target diastolic point.

\subsubsection{Cube-based sort}
The representation form of point clouds is vectors in three-dimensional space. When the points are stored in an orderly manner, the estimation of the deformation field can be obtained as follows:
\begin{equation}\label{Dc}
	\mathcal{D}_c = C_d - C_s
\end{equation}

Attempting to use vector subtraction to obtain the deformation field is incorrect, and the point cloud should be sorted first. The extracted part of the vascular centerline ($C_c$) calculates the span on the x, y, and z axes. The span ($\mathbf{s}$) is defined as follows:
\begin{equation}
	\mathbf{s}_i = \|\mathcal{M}(C_c^i) - \mathcal{L}(C_c^i) \|_1, i\in \{x,y,z\}
\end{equation}
where $\mathcal{M}$ is the maximum value function, and $ \mathcal{L}$ denotes the minimum. Span is used to characterize the trend of a 3D line on the coordinate axis. The extracted centerline can be sorted along the trend. Specifically, as follows:
\begin{equation}
	C_{\text{sorted}} = \mathcal{S}(C_c, a), \mathbf{s}_{a} = \mathcal{M}(\mathbf{s}_i)
\end{equation}
where, $C_{\text{sorted}}$ represents the sorted point cloud, and $\mathcal{S}$ denotes the sorting function. $a$ is the trend axis. After sorting the centerlines of the systole and diastole, the deformation field $\mathcal{D}_c$ is derived by Eq. \ref{Dc}.

\subsection{Vascular deformation and interpolation}
As described in Eq. \ref{R}, the registration of vessels depends on the point by point deformation vector. The displacement of the coronary artery in the body is small, and the displacement vectors of adjacent vessel points are similar to Fig. \ref{Method} (C), clustering vessel points using the centerline assumes that the clustered points have the same displacement vector as the centerline points. By slicing coronary artery points, the displacement vectors of each part can be derived. We propose a cuboid-based points extraction that utilizes the coordinates of the centerline for clustering and obtaining slices. Specifically, the cuboid is defined as follows:
\begin{equation}
	\mathcal{B} = Cuboid(\mathbf{o}, \mathbf{l}, \mathbf{w}, \mathbf{h})
\end{equation}
where, $\mathcal{B}$ represents the region for extracting vascular points based on centerline points. $Cuboid$ is the function that defines the region. $\mathbf{o}$ is the center point of the cube, i.e. the centerline point. $\mathbf{l}$ denotes the length, set to 1. $\mathbf{w}$ is the width, equal to the value of $\mathbf{l}$. $\mathbf{h}$ represents the height of the cuboid, determined by the distance between adjacent centerline points, such as, $\mathbf{h} = \| c_i - c_{i-1}\|$, $c_i \in C_c$.

The deformation field of vessel points is calculated as follows:
\begin{gather}
	V_{s}^{\Sigma} = \mathcal{B}(V_s, x), x\in C_s \\
	\mathcal{D}_i = \mathcal{D}_{cx}, i\in V_{s}^{\Sigma}
\end{gather}
where $V_{s}^{\Sigma}$ represents the extracted vascular point set. $\mathcal{D}_{cx}$ is the deformation vector of the centerline point $x$, and $\mathcal{D}_i$ denotes the point in set $V_{s}^{\Sigma}$. Finally, by interpolating the deformation field $\mathcal{D}_i$ of the vessel points, the path vector $\mathcal{D}_t$ at any time can be obtained. According to Eq. \ref{R}, coronary arteries at any phases can be obtained through systole and $\mathcal{D}_t$.

\section{Experiments}
\subsection{Dataset}
Clinical data is collected from 54 patients in the Department of Cardiothoracic Surgery at Renji Hospital in Shanghai, China. For each patient, obtain CTA volumes for 4 different phases ($30\%, 45\%, 60\%, 75\%$), with approximately 300 slices. 10 patients with clearer intermediate phase are selected as the test-set, and the 44 patients are selected as the training-set. In this paper, the values of HD and CD are multiplied by $10^{2}$ for comparison purposes.

\subsection{Synthetic data}
Covering the sample space as much as possible can effectively improve the accuracy of the model. For paired vascular point clouds, it is difficult to perform data synthesis. Using the centerline to represent coronary arteries not only reduces the difficulty of prediction, but also facilitates sample synthesis. Bezier curve is used to obtain a set of points in three-dimensional space. Modify the length by taking different sizes.

\begin{table}[t]
	\renewcommand\arraystretch{1.2}
	\caption{The interpolation errors of coronary artery trees varies with the proportion of invisible vessels. }
	\centering
	\resizebox{\linewidth}{!}{
		\begin{tabular}{c|c|ccc|ccc|cc}
			\hline
			\multirow{2}*{ Proportion }&{  }&{}&{Overall}&{}&{}&{Partial}&{}&{} &{}\\
			~&{  }&{DCP}&{FMR}&{ICP}&{DCP}&{FMR}&{ICP}&{Ours (DL)}&{Ours (Sort)} \\
			\hline
			\hline
			\multirow{2}*{  $3.17\%$}&{CD}&{2.79}&{1.49}&{0.48}&{0.60}&{0.45}&{0.23}&{0.021} &{0.037}\\
			{     ~     }&{HD}&{14.93}&{12.61}&{6.63}&{20.48}&{8.176}&{5.12}&{0.143} &{0.147}\\
			\hline
			\multirow{2}*{  $29.11\%$}&{CD}&{1.72}&{2.65}&{0.59}&{0.91}&{0.34}&{0.42}&{0.015} &{0.016}\\
			{      ~     }&{HD}&{15.08}&{19.80}&{6.48}&{16.99}&{8.07}&{5.04}&{0.102} &{0.14}\\
			\hline
			\multirow{2}*{  $58.77\%$}&{CD}&{20.94}&{2.25}&{0.25}&{1.22}&{0.23}&{0.21}&{0.053} &{0.050}\\
			{      ~     }&{HD}&{22.17}&{33.97}&{7.97}&{19.29}&{4.65}&{3.36}&{0.11} &{0.12}\\
			\hline
	\end{tabular}}
	\label{Vascular}
\end{table}

\begin{figure}[t]
	\centering
	\begin{minipage}[t]{0.56\linewidth}
		\centering
		\includegraphics[width=\linewidth]{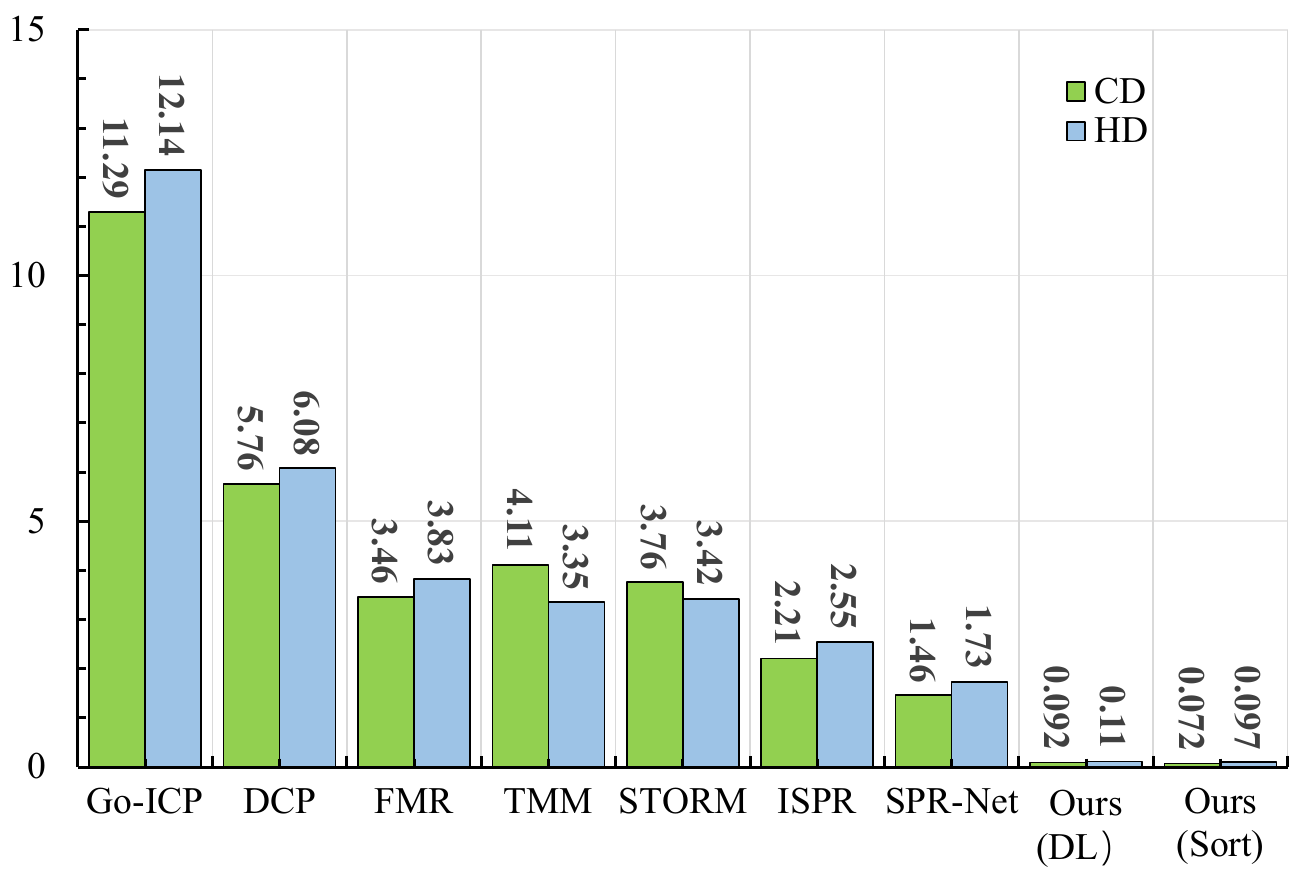}
		\caption{Comparison with state-of-the-art methods for registering point clouds during systole and diastole.}
		\label{CD_HD}
	\end{minipage} \quad
	\begin{minipage}[t]{0.38\linewidth}
		\centering
		\includegraphics[width=\linewidth]{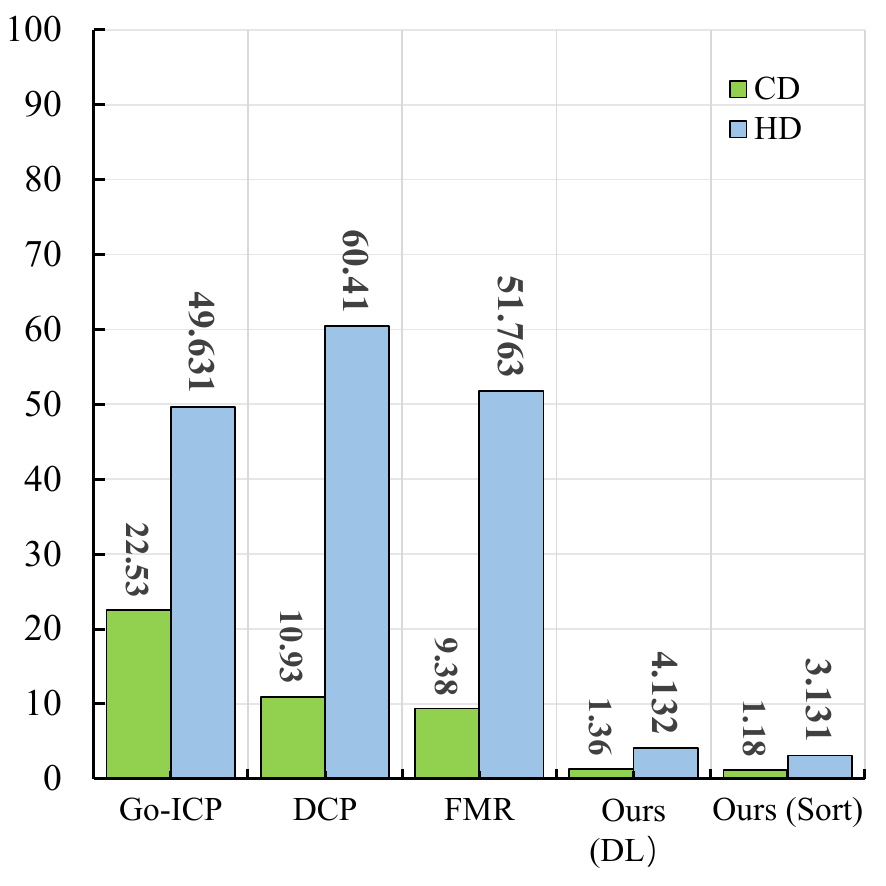}
		\caption{Comparison results of interpolation.}
		\label{InterpolationCompare}
	\end{minipage}
\end{figure}

\begin{figure}[t]
	\centerline{\includegraphics[width=\linewidth, scale=1.00]{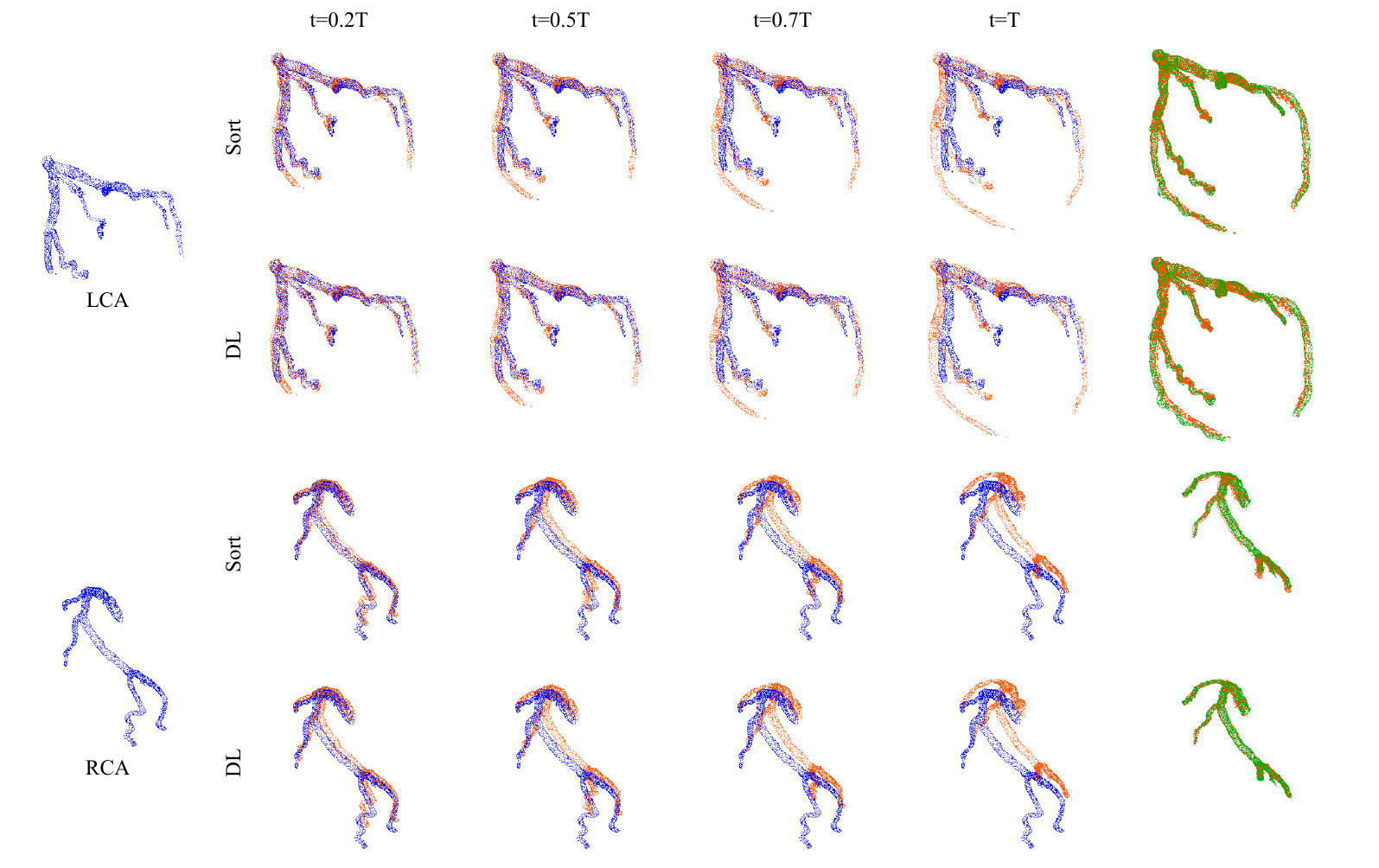}}
	\caption{Visualization results of interpolation. Blue is the systole, green represents the diastole, and orange denotes the interpolation result.}
	\label{Interpolation}
\end{figure}
\begin{table}[t]
	\renewcommand\arraystretch{1.2}
	\caption{Ablation study. }
	\centering
	\resizebox{\linewidth}{!}{
		\begin{tabular}{c|c|cccccc}
			\hline
			&&\multicolumn{2}{c}{CA00}&\multicolumn{2}{c}{CA01}&\multicolumn{2}{c}{CA02}\\
			&&{CD}&{HD}&{CD}&{HD}&{CD}&{HD}\\
			\hline
			\hline
			{}&{w.o. Outdegree}				 	&{0.0741}&{0.3811}&{0.1362}&{0.3920}&{0.1541}&{0.2252}\\
			{Sort}&{w.o. Sort}	&{11.617}&{52.431}&{31.379}&{83.550}&{9.4672}&{102.242}\\
			{}&{Ours}	&{0.0665}&{0.1801}&{0.1181}&{0.2331}&{0.1064}&{0.2214}\\
			\hline
			{}&{w.o. Outdegree}	&{0.0699}&{0.2533}&{0.1115}&{0.2132}&{0.1143}&{0.2401}\\
			{DL}&{w.o. soft-DTW}						&{0.1409}&{0.2431}&{0.1571}&{0.4317}&{0.1341}&{0.2121}\\
			{}&{Ours}				&{0.0678}&{0.1622}&{0.0987}&{0.1930}&{0.0921}&{0.1871}\\
			\hline
	\end{tabular}}
	\label{ablation}
\end{table}

\subsection{Results}

\textbf{Registration comparison:} Some point cloud registration methods are compared, including Go-ICP \cite{yang2015go}, DCP \cite{wang2019deep}, FMR \cite{huang2020feature}, TMM \cite{ccimen2016reconstruction}, STORM \cite{wang2022storm}, ISPR \cite{chen2020unsupervised}, and SPR-Net \cite{zhang2023spr}. SPR-Net \cite{zhang2023spr} is a framework specifically designed for coronary points registration during systole and diastole. The comparison results are shown in Fig. \ref{CD_HD}. The two proposed methods, based on path wise point-to-point mapping, result in better registration results.

\textbf{Interpolation of vascular deficiency:} Due to the lack of large-scale clinical datasets, deep learning based point cloud interpolation methods cannot be compared. Therefore, in order to verify the effectiveness of 4D-CAT, we propose three baseline methods, which are DCP \cite{wang2019deep}, FMR \cite{huang2020feature}, and ICP. To simulate non visualization of blood vessels, certain branch deletions are manually set on clinical data. The results are shown in Table \ref{Vascular}. Our method still demonstrates excellent performance, although it is also affected by the lack of vascular imaging.

\textbf{Interpolation comparison:} The dataset used in this paper contains four different phases, so the coronaries with $45\%$ and $60\%$ phases are used as the ground-truth for validation. We chose Go-ICP \cite{yang2015go}, DCP \cite{wang2019deep}, and FMR \cite{huang2020feature} for comparison. The comparison results are shown in Fig. \ref{InterpolationCompare}. 4D-CAT interpolation results are more accurate and have fewer outliers.

\textbf{Interpolation visualization:} The visualization result is shown in Fig. \ref{Interpolation}, which includes three periods interpolated in the middle. T, as a cardiac cycle, represents the exercise time of the coronary artery from systole to diastole. The proposed method can effectively demonstrate the deformation process of blood vessels for LCA and RCA. The last column displays the overlap between the mapped systolic points and diastole (green), verifying the accuracy of path prediction.

\textbf{Ablation study:} As shown in Table \ref{ablation}, two methods are used to conduct ablation study separately for three objects. In the sort-based method, we explored the importance of outdegree and sort. Firstly, following the method proposed in \cite{wu2022car}, identify bifurcation points and segment coronary arteries based on voxel and neighborhood search. According to the results, the proposed strategies in both methods are effective and have other potential application scenarios.

\section{Conclusion}
In this paper, we are the first to propose a scheme for generating 4D coronary artery trees from systolic and diastolic coronary point clouds. Our method is largely superior to state-of-the-art methods and also achieves interpolatable intermediate processes. This indicates the potential applicability of 4D-CAT in digital organs and real-world clinical scenarios.


\end{document}